\documentclass{article}

\usepackage{microtype}
\usepackage{graphicx}
\usepackage{subcaption}
\usepackage{booktabs} 

\usepackage{hyperref}

\usepackage[preprint]{icml2026}

\usepackage{amsmath}
\usepackage{amssymb}
\usepackage{mathtools}
\usepackage{amsthm}

\usepackage[capitalize,noabbrev]{cleveref}

\usepackage{enumitem}
\usepackage{multirow}

\theoremstyle{plain}

\theoremstyle{definition}

\theoremstyle{remark}

\usepackage[textsize=tiny]{todonotes}
\usepackage{fontawesome}

\icmltitlerunning{LocateEdit-Bench: A Benchmark for Instruction-Based Editing Localization}

\begin{document}

\twocolumn[
  \icmltitle{LocateEdit-Bench: A Benchmark for Instruction-Based Editing Localization}



  \icmlsetsymbol{equal}{*}
  \icmlsetsymbol{corr}{\faEnvelopeO}

  \begin{icmlauthorlist}
    \icmlauthor{Shiyu Wu}{equal,casia,baai,ucas}
    \icmlauthor{Shuyan Li}{equal,queen}
    \icmlauthor{Jing Li}{harbin}
    \icmlauthor{Jing Liu}{casia,ucas}
    \icmlauthor{Yequan Wang}{corr,baai,peking}
  \end{icmlauthorlist}

  \icmlaffiliation{casia}{Institute of Automation, Chinese Academy of Sciences, Beijing, China}
  \icmlaffiliation{baai}{Beijing Academy of Artificial Intelligence, Beijing, China}
  \icmlaffiliation{ucas}{University of Chinese Academy of Sciences, Beijing, China}
  \icmlaffiliation{queen}{School of EEECS, Queen's University Belfast, Northern Ireland, UK}
  \icmlaffiliation{harbin}{Harbin Institute of Technology, Shenzhen, China}
  \icmlaffiliation{peking}{Peking University, Beijing, China}
  
  \icmlcorrespondingauthor{Yequan Wang}{tshwangyequan@gmail.com}

  \icmlkeywords{Forgery Localization, Manipulation Localization, Machine Learning}

  \vskip 0.3in
]



\printAffiliationsAndNotice{\icmlEqualContribution}


\begin{abstract}

Recent advancements in image editing have enabled highly controllable and semantically-aware alteration of visual content, posing unprecedented challenges to manipulation localization. 
However, existing AI-generated forgery localization methods primarily focus on inpainting-based manipulations, making them ineffective against the latest instruction-based editing paradigms. 
To bridge this critical gap, we propose \textbf{LocateEdit-Bench}, a large-scale dataset comprising $231$K edited images, designed specifically to benchmark localization methods against instruction-driven image editing. 
Our dataset incorporates four cutting-edge editing models and covers three common edit types. 
We conduct a detailed analysis of the dataset and develop two multi-metric evaluation protocols to assess existing localization methods. 
Our work establishes a foundation to keep pace with the evolving landscape of image editing, thereby facilitating the development of effective methods for future forgery localization. 
Dataset will be open-sourced upon acceptance. 

\end{abstract}
    
\section{Introduction}

The increasing photorealism and accessibility of image manipulation technologies pose a growing threat to the integrity of digital media. 
To uphold trust in visual information, precisely localizing tampered regions within an image has become a critical task. 
This challenge is significantly amplified by the advent of AI-powered generative models \citep{gan17, ddpm1}, which enable highly realistic and semantically coherent edits that are far more subtle and difficult to detect than those created by traditional methods like splicing and copy-move \citep{gre2}. 
Consequently, there is an urgent need to develop robust localization techniques capable of keeping pace with these rapidly evolving forgery methods. 


\begin{figure}[t]
  \vskip 0.2in
  \centering
  \subfloat[Inpainting pipeline]{
        \includegraphics[width=0.96\linewidth]{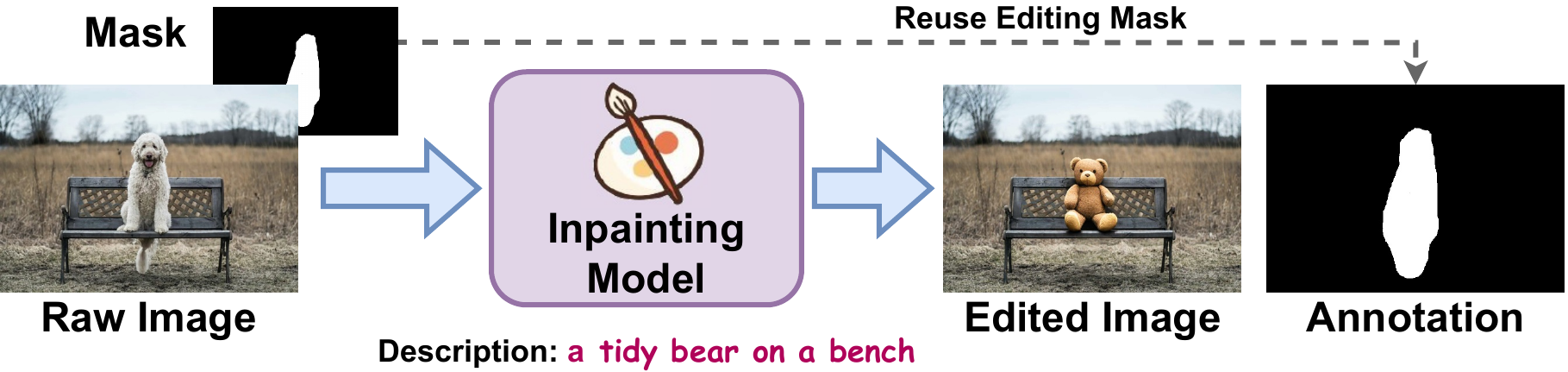}
        \label{fig:inpainting}
    }
    \hfill
    \subfloat[Instruction-based editing pipeline]{
        \includegraphics[width=0.96\linewidth]{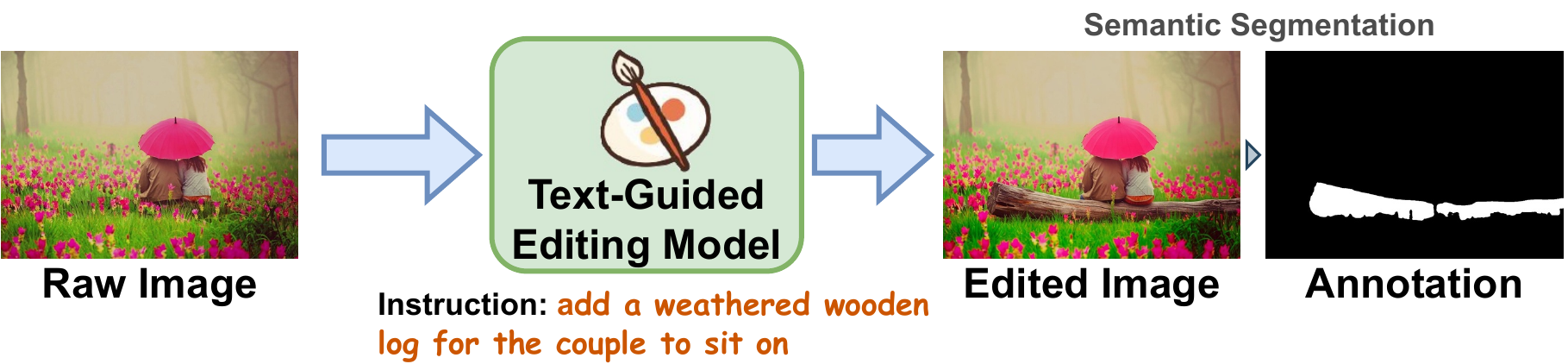}
        \label{fig:textguided}
    }
  \caption{Comparison of two datasets using different editing approaches. Localizing edits in instruction-based image editing is particularly difficult because the edits are semantically coherent and visually seamless.}
  \label{fig:editmethods}
  \vskip -0.1in
\end{figure}

Historically, inpainting with diffusion models~\citep{repaint3} has been a predominant approach for content-aware manipulation.
As illustrated in \cref{fig:inpainting}, it relies on a user-provided mask to constrain edits to a specific region. 
The explicit mask inherently provides a strong, low-level boundary signal between authentic and altered content, leading to detectable artifacts or statistical mismatches which simplify the localization task without the need to assess deeper semantic coherence. 
However, the field is rapidly shifting with the emergence of instruction-based editing models. 
These tools, such as Multi-Modal Diffusion Transformers (MMDiT) \citep{sd3_5}, now enable users to manipulate images through natural language alone \citep{showo40, step1xedit7}. 
As depicted in \cref{fig:textguided}, instruction-based editing models produce semantically coherent and visually seamless edits without any predefined mask. 
While this flexible and realistic paradigm has become the de facto standard in contemporary image editing tools, it remains susceptible to going undetected due to the absence of dedicated localization methods. 

To address this critical gap, we present LocateEdit-Bench, the first large-scale benchmark dataset specifically designed to systematically investigate and benchmark the forgery localization challenge posed by modern instruction-based image editing models. 
LocateEdit-Bench comprises $231$K edited images, generated by $4$ state-of-the-art instruction-based editors. 
To ensure representational diversity and practical relevance, our dataset encompasses $3$ fundamental object-level manipulation types---object addition, replacement, and attribute modification---applied to high-quality natural images. 
Images edited by high-consistency models are subsequently processed by a high-quality semantic segmentation model to generate precise masks. 
During our construction process, we employ both Large Language Models (LLM) and Vision-Language Models (VLM) to jointly filter instructions and images, resulting in a high-quality and comprehensive dataset. 

Building upon this dataset, we conduct a comprehensive benchmark evaluation of state-of-the-art localization methods. To enable a thorough assessment, we design a multi-faceted evaluation protocol that includes a cross-generator generalization test, examining how well methods perform across different editing models. 
Our extensive experiments reveal significant performance drops and inherent limitations of existing techniques when faced with instruction-based edits, providing crucial insights into the unique challenges of this emerging paradigm. 
We believe LocateEdit-Bench establishes an essential foundation for developing and evaluating robust localization methods, thereby helping the research community keep pace with the evolving landscape of AI-powered image manipulation. 

Our contributions are summarized as follows: 
\begin{itemize}[topsep=0pt, itemsep=0pt]
    \item We systematically investigate the task of instruction-based image editing localization, a critical step in adapting forensic methods to the current paradigm of image manipulation. 
    
    \item We introduce LocateEdit-Bench, the first large-scale benchmark dedicated to localizing modern instruction-based image edits, featuring $231$K samples from $4$ editors across $3$ manipulation types. 
    
     \item We perform an extensive benchmark evaluation of existing state-of-the-art localization methods, uncovering their critical limitations on this challenging task.
\end{itemize}

\section{Related Work}

\begin{figure*}[t]
    \vskip 0.2in
    \centering
    \includegraphics[width=1.0\linewidth]{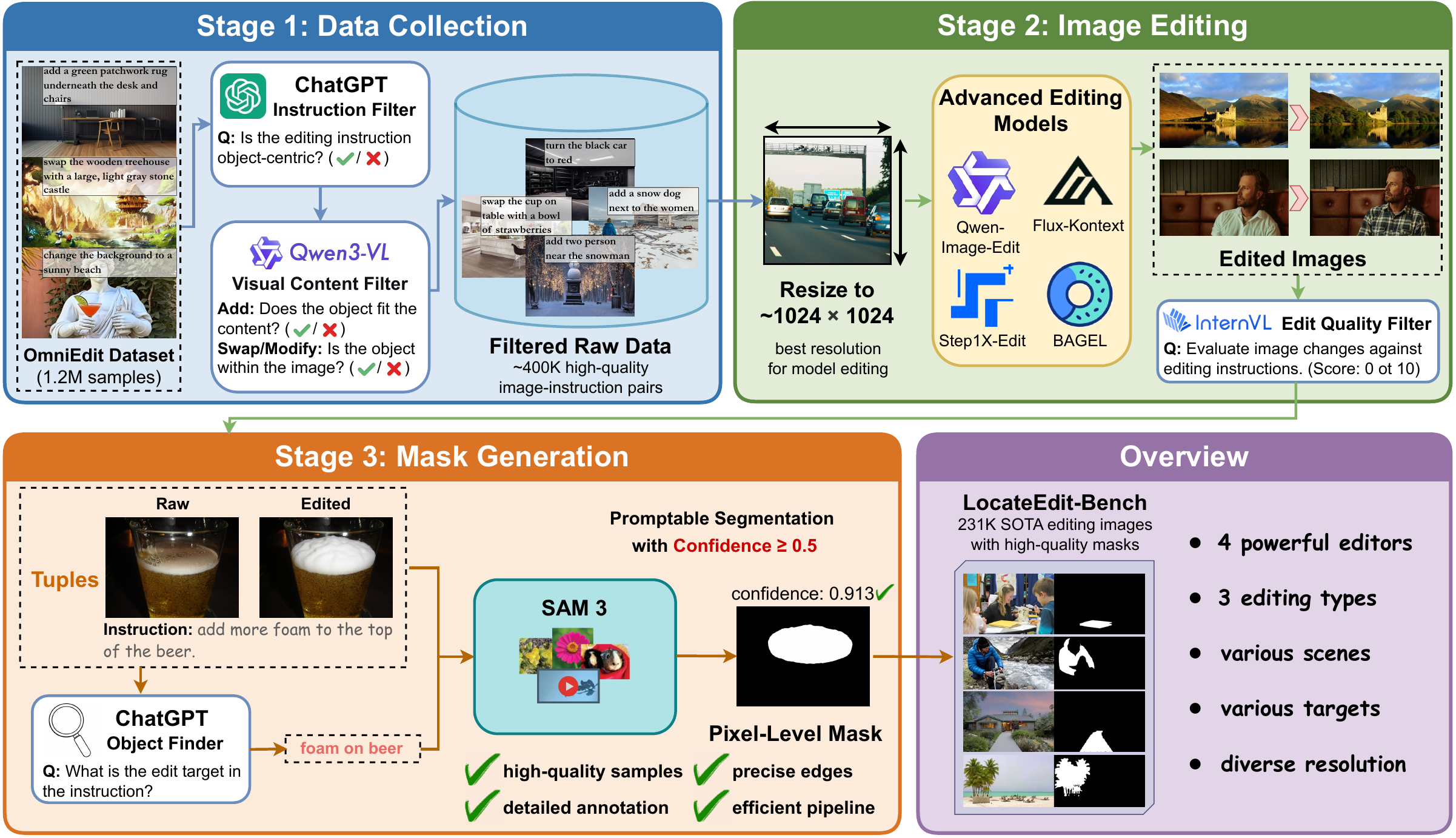}
    \caption{Construction pipeline of LocateEdit-Bench. We carefully select suitable real-world images and editing prompts from a large-scale editing dataset. Then we employ four latest image editing models to generate $231$K high-quality edited images. Subsequently, precise masks are generated using a high-quality semantic segmentation model. }
    \label{fig:pipeline}
\end{figure*}

\subsection{Image Editing}

Diffusion Models \citep{ddpm1} have made notable breakthroughs in image editing by offering enhanced controllability. 
Early studies such as DiffusionCLIP \citep{diffusionclip9} and P2P \citep{p2p8} enable semantic and stylistic modifications by leveraging cross-attention mechanisms. 
Subsequently, instruction-driven models like InstructPix2Pix \citep{pix2pix10} further advanced the field by allowing edits to be guided through plain text. 
However, these methods often struggle to achieve precise and localized edits, frequently introducing artifacts or inconsistencies in non-target regions. 
In this context, inpainting distinguishes itself by targeting the restoration of missing regions defined by an explicit mask, which inherently promotes higher editing consistency. 
Moreover, training-free inpainting methods \citep{repaint3} can rapidly adapt to newly released diffusion models, harnessing their enhanced generative capabilities to produce superior editing results. 
The high fidelity and clear manipulation boundaries make inpainting central to forgery localization research. 

Recently, the advent of MMDiT has catalyzed the development of instruction-based image editing models, far surpassing the performance of prior approaches. 
Leveraging strong language-vision alignment and ample synthetic training data, models such as BAGEL \citep{bagel11} and Gemini $2.5$ \citep{gemini12} achieve remarkable editing fidelity and flexibility using natural language alone. 
This advancement not only enables edited content to seamlessly blend into original images but also supports a wide range of complex editing tasks, leading to its broad adoption. 
Consequently, the lack of discernible traces pose significant new challenges for forgery detection and localization.

\subsection{Image Manipulation Datasets}
Early image manipulation datasets \citep{casia13, web14} are primarily constructed using traditional image processing techniques, such as splicing and copy-move operations. 
With the advancement of deepfake technologies, subsequent works like IMD2020 \citep{imd2020_15} and AutoSplice \citep{autosplice16} begin to incorporate images edited by GANs \citep{gan17} or diffusion models. 
Currently, most research \citep{tgif18, cocoinpainting19} focuses on inpainting-based edits, where the manipulated regions are readily defined by pre-existing masks within the editing process. 
Although these datasets \citep{gim4} have served as important foundations for forgery localization research, they are inherently limited by the performance bottlenecks of the editing models, which confines edits to a narrow target set and compromises consistency in non-edited regions. 
These inherent biases can thereby negatively impact the performance of localization methods. 

While recent MMDiT-based editing models \citep{qwenimage6, bagel11} excel at background consistency, they also render the edited target increasingly difficult to identify. 
To the best of our knowledge, the robustness of existing localization methods against such advanced, instruction-based manipulations remains largely unexplored. 
Our work aims to bridge this gap by introducing a dedicated benchmark for this emerging field, thereby extending the study of image forensics into the new era of modern instruction-based generative editing. 
\section{LocateEdit-Bench Dataset}

This section introduces LocateEdit-Bench, a comprehensive manipulation localization dataset which comprises a total of $231$K high-quality, high-resolution edited images, each accompanied by a corresponding segmentation mask. 
The edited images are generated by four distinct editing models, with each model contributing at least $50$K samples to ensure a balanced distribution across the dataset. 
In the following sections, we detail how to construct LocateEdit-Bench, for which the full pipeline is illustrated in \cref{fig:pipeline}. 

\begin{figure*}[t]
    \centering
    \includegraphics[width=0.95\linewidth]{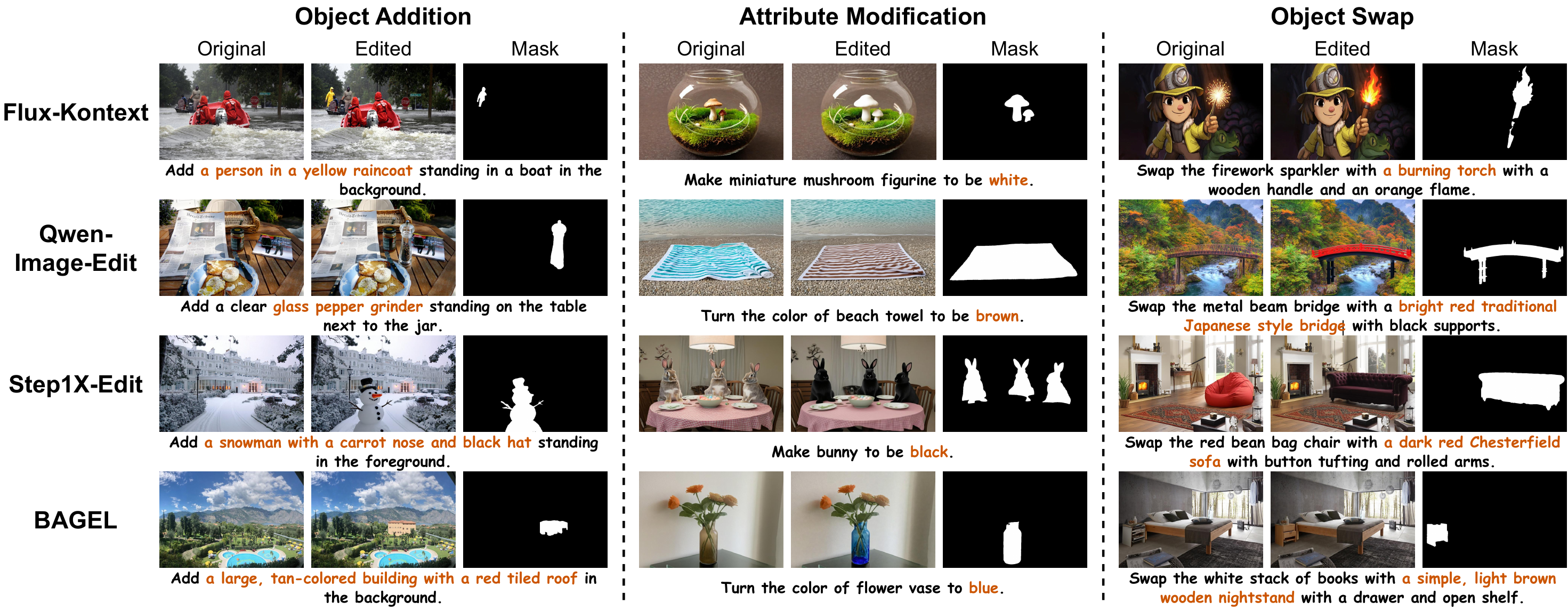}
    \caption{Samples of LocateEdit-Bench. LocateEdit-Bench constitutes a comprehensive benchmark for editing localization, featuring high-resolution images edited by four different models, and containing three edit types applied to targets of diverse sizes. }
    \label{fig:samples}
\end{figure*}

\subsection{Images and Instructions Collection}

In this work, we adopt a subset of real images from the OmniEdit \citep{omniedit23} dataset to construct our dataset. 
The original images in OmniEdit are sourced from LAION \citep{laion400m24} and OpenImages \citep{openimages25}, which provide a diverse collection of images spanning a wide range of categories, rich content, and varying resolutions. 
Each image is associated with several coherent editing instructions generated by GPT-4o, covering a variety of editing scenarios such as object manipulation, style transfer, and background replacement. 
We begin by filtering the raw images and their corresponding instructions based on key criteria, primarily focusing on the separability of the main editing subject and the appropriateness of the edit target for the scene. 
Based on the analysis of instruction-based editing characteristics, we select three editing types---object swap, object addition, and attribute modification---as the focus of our study. 
To enforce semantic plausibility in the edits, we use Qwen3-VL \citep{qwen3vl26} to screen instructions, discarding those where the target object would be contextually incongruous (for addition) or is not originally present in the image (for swap and modification). 
After this process, approximately $400$K image-instruction pairs are retained from the original $1.2$M samples in OmniEdit. 

\subsection{Image Editing}

While the OmniEdit dataset provides abundant edited images, it relies on outdated editing methods such as Prompt-to-Prompt (P2P) \citep{p2p8}, whose performance is limited and often fails to achieve precise, targeted edits. Thus, we regenerate the edited images using the original real images and their corresponding editing instructions, but with more powerful editing models. We employ four open-source state-of-the-art editing models, including Qwen-Image-Edit \citep{qwenimage6}, Flux-Kontext \citep{fluxkontext27}, Step1X-Edit \citep{step1xedit7}, and BAGEL \citep{bagel28}, which represent both specialized editing architectures based on MMDiT and unified generation-understanding frameworks. 
These modern models deliver significantly improved directional control and background consistency, resulting in edits that are both accurate and coherent. 

Building upon the four aforementioned editors, we process the collected images according to a structured pipeline. 
For each editing type, the corresponding data is evenly divided into four subsets, each of which is then edited by a designated model. 
To strike a balance between model compatibility and dataset diversity, all original images are resized to approximately $1024 \times 1024$ pixels. 
This operation not only meets the preference of editing models for optimal performance but also enhances the resolution diversity within the dataset. 
Other key inference parameters, such as guidance scale and sampling steps, are set in accordance with the recommendations for each specific model for the purpose of obtaining the best possible editing outcome. 
Upon completion of the generation process, we follow \citet{omniedit23} and additionally employ InternVL2 \citep{internvl2_29} to evaluate editing accuracy and background consistency, thereby filtering out low-confidence samples. 
This controlled generation process allows us to systematically produce high quality editing results, yielding a reliable dataset for subsequent research.

\subsection{Mask Generation}

Since instruction-based editing does not inherently provide segmentation boundaries, a core challenge is to precisely define the region that has been edited. 
In this context, our prior focus on object-centric editing targets aligns well with the capacity of segmentation model for precise region delineation, which is consistent with human perception of the edited area. 
To generate the mask of the target object for editing, we adopt a comprehensive multi-model segmentation framework. 
First, We utilize ChatGPT to analyze the target by processing carefully designed prompt templates, which help identify the object to be edited while preserving certain modifiers to specify its exact instance in the image. 
Subsequently, these textual cues are fed into SAM3 \citep{sam3_39}, a powerful promptable segmentation model which is able to accurately segment the specified target from the image. 
In addition to the predicted pixel-level mask, SAM3 provides a confidence score for each segmentation. 
We leverage this by applying a confidence threshold of above $0.5$ to filter out unreliable results, ensuring the quality of the segmentation output. 

\subsection{Analyses of LocateEdit-Bench}

\begin{figure}[t]
  \centering
  \subfloat[Category Distribution]{
        \includegraphics[width=0.45\linewidth]{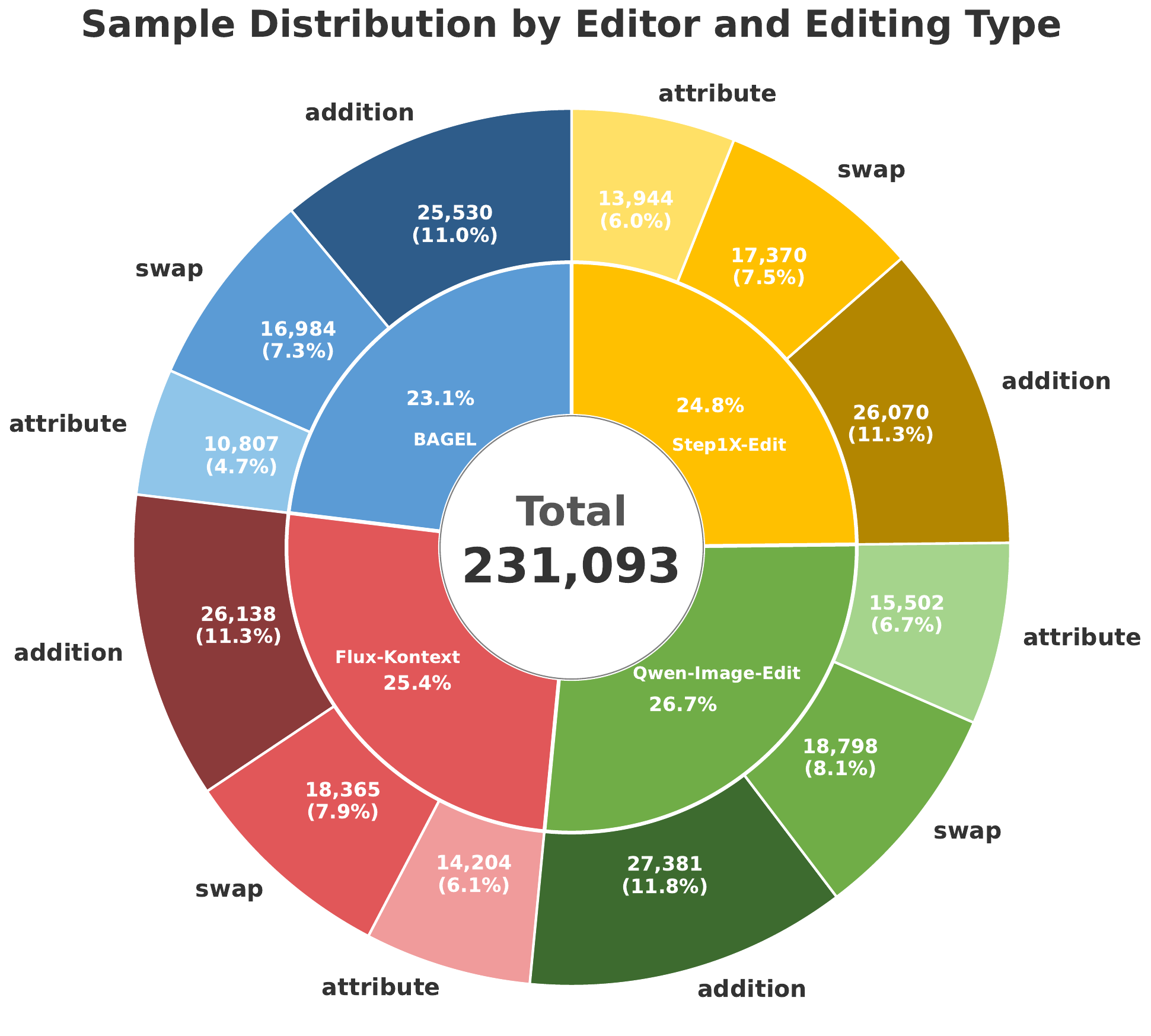}
        \label{fig:datadistribution}
    }
    \subfloat[Instruction Word Cloud]{
        \includegraphics[width=0.40\linewidth]{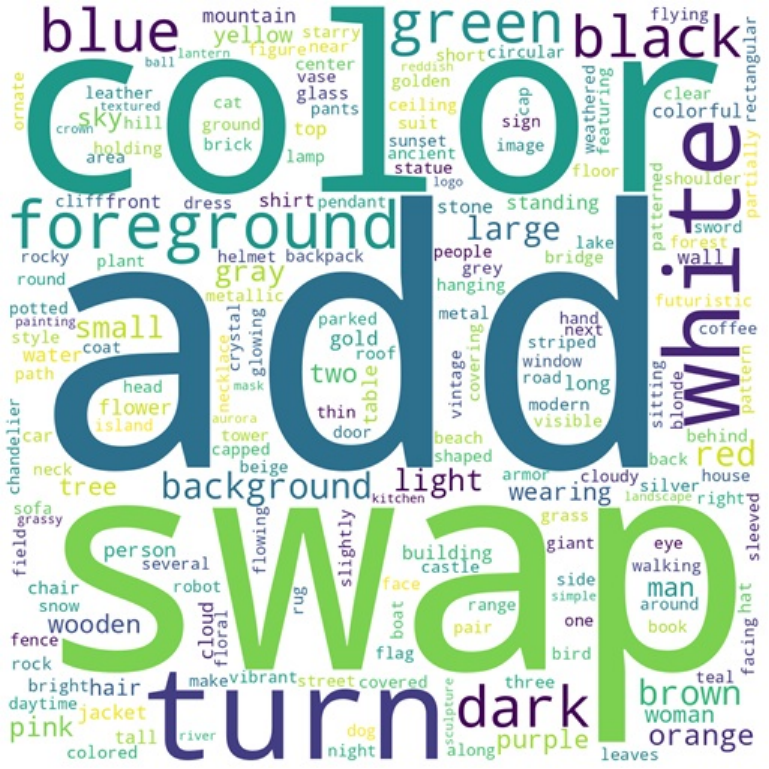}
        \label{fig:wordcloud}
    }
  \caption{Categroy distribution of LocateEdit-Bench and word cloud of editing instructions. }
  \label{fig:statistic}
  \vskip -0.2in
\end{figure}

\begin{figure}[t]
  \vskip 0.2in
  \centering
  \subfloat[Edited Regions]{
        \includegraphics[width=0.45\linewidth]{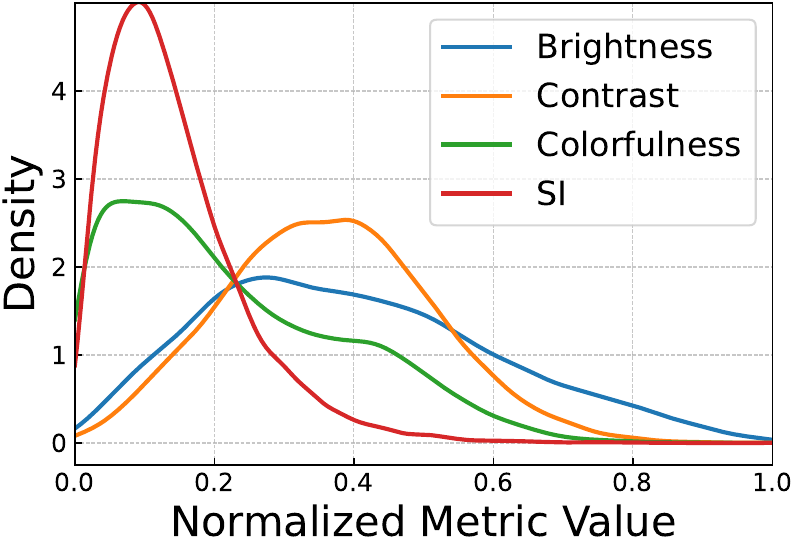}
        \label{fig:flawsa}
    }
    \subfloat[Background]{
        \includegraphics[width=0.45\linewidth]{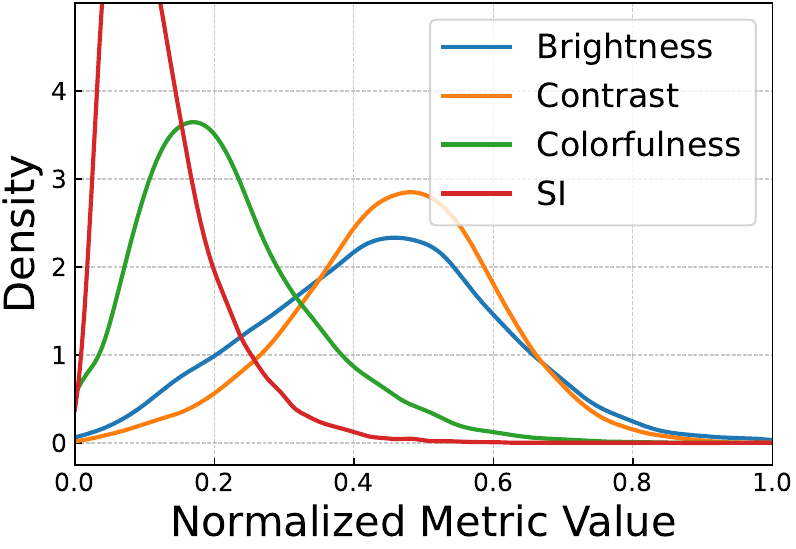}
        \label{fig:flawsb}
    }
  \caption{Comparison of feature distributions in edited regions and background areas. Edited regions show significantly reduced colorfulness and brightness, yet exhibit slightly increased spatial information (SI). }
  \label{fig:featuredistribution}
  \vskip -0.1in
\end{figure}

Ultimately, $231$K edited images with detailed annotations are selected to constitute the final dataset, formally establishing the foundation for our editing localization research. 
Representative samples are illustrated in \cref{fig:samples}. 
The category distribution is presented in \cref{fig:datadistribution}, and a word cloud depicting the editing instructions is shown in \cref{fig:wordcloud}.
Collectively, these visualizations demonstrate the diversity of editing targets and categories within our dataset. 
The scale and quality of of it are sufficient to support stable model convergence during training. 
Upon building the dataset for instruction-based editing localization, a critical question must be addressed: is the edited content truly distinguishable from its authentic surroundings?
Empirically, since the edited content originates from textual instructions and the surrounding context remains constrained by the original image, a measurable cross-modal discrepancy is thus introduced, enabling their potential separation. 
To validate this proposition, we analyze images edited by Qwen-Image-Edit \citep{qwenimage6} and compute the feature distributions of the foreground edited areas and the background. 
As shown in \cref{fig:featuredistribution}, several distinct patterns emerge from the feature comparison. Specifically, edited regions exhibit significantly reduced colorfulness (green line) and brightness (blue line), while showing a slight increase in spatial information (red line). 
This observation indicates that edited content carries a detectable signature distinct from its context, confirming that localizing instruction-edited objects is both reasonable and feasible, thereby firmly supporting the validity of our proposed task. 

LocateEdit-Bench provides a rich, diverse foundation for systematically investigating manipulation localization on modern editing models. 
We expect this dataset to contribute to advances in image manipulation detection and localization research, and thereby to a more nuanced understanding of present-day editing traces. 

\section{Experiment}

In this section, we conduct a comprehensive evaluation of existing methods on our proposed LocateEdit-Bench. We begin by detailing the experimental settings, including evaluation protocols and metrics. Then we introduce the selected baseline methods for localization task and analyze the results under multiple protocols.

\subsection{Evaluation Protocols}

Our dataset comprises editing data from four state-of-the-art editing models, each with its own separate training and testing sets. 
To build the test set, we adopt a stratified sampling strategy to maintain the distribution of editing types. 
Specifically, a total of $6$K images are selected from the outputs of each editing model, proportionally representing its internal composition of edit types. 
This results in a balanced and model-specific test subset of $24$K images in total. 
To thoroughly assess model performance and generalization, we design two primary evaluation protocols: 

\begin{itemize}[topsep=0pt, itemsep=0pt, leftmargin=10pt]
\item \textbf{Protocol 1: Full-Set Evaluation.} Models are trained on a composite dataset containing images generated by all editing model categories and tested on each individual category. 
This protocol evaluates the capability to localize diverse types of edits. 

\item \textbf{Protocol 2: Cross-Editor Generalization.} Models are trained exclusively on data from a single editing model category and tested on all categories. This protocol rigorously tests the ability to generalize to novel, unseen manipulation techniques, which is crucial for real-world applications. 
\end{itemize}

Additionally, we evaluate all models under various image degradations, such as JPEG compression and Guassian blurring, to examine their practical robustness against common post-processing operations. 

\subsection{Metrics}

Since our research focuses on the localization of manipulated regions, we employ pixel-level metrics for evaluation. 
Following established practices \citep{cocoinpainting19} in the field, we compute Accuracy, Area Under the Curve (AUC), and F1-Score at the pixel level, where each pixel is treated as an instance in a binary classification task. 
A threshold of $0.5$ is applied to most models for calculating Accuracy and F1-Score, while for ObjectFormer, which adopts the Equal Error Rate (EER) threshold, we compute the corresponding threshold separately. 
Furthermore, to ensure a more comprehensive and robust assessment of spatial overlap between predictions and ground truth, we incorporate the Dice Similarity Coefficient (Dice Score) and Intersection over Union (IoU) as complementary metrics. 
These metrics collectively provide a thorough assessment of both classification correctness and region segmentation quality. 

\begin{table*}[t]
    \vskip 0.2in
    \centering
    \caption{Pixel-level evaluation results following Protocol 1. The Accuracy and F1-Score are computed over all pixels, while the AUC represents the average of image-wise AUC scores. The best results are highlighted in \textbf{bold}, and the second-best are \underline{underlined}.}
    \label{tab:cls}
    
    \begin{tabular}{l|c|c|c|c|c}
        \toprule
        \multirow{2}{*}{Method} & Qwen-Image-Edit & Flux-Kontext & Step1X-Edit & BAGEL & Average \\
        \cline{2-6}
         & ACC / AUC / F1 & ACC / AUC / F1 & ACC / AUC / F1 & ACC / AUC / F1 & ACC / AUC / F1 \\
        \hline
        ObjectFormer & 94.1 / 95.0 / 79.3 & 95.4 / 96.9 / 83.7 & 93.7 / 94.7 / 76.7 & 93.7 / 95.2 / 77.5 & 94.2 / 95.5 / 79.3 \\
        PSCC-Net & 91.0 / 95.3 / 75.4 & 93.0 / 96.5 / 79.9 & 89.3 / 93.7 / 71.3 & 90.8 / 95.0 / 75.4 & 91.0 / 95.1 / 75.5 \\
        IML-ViT & 93.5 / 94.9 / 78.9 & 95.5 / 97.2 / 85.3 & 93.4 / 94.5 / 78.0 & 94.2 / 96.1 / 80.9 & 94.1 / 95.7 / 80.8 \\
        PIM-Net & 93.6 / 95.9 / 80.9 & 95.5 / 97.0 / 85.9 & 92.6 / 94.6 / 77.9 & 94.6 / 96.2 / 83.3 & 94.1 / 95.9 / 82.0 \\
        Mesorch & \textbf{96.1} / \textbf{97.5} / \underline{86.9} & \underline{96.5} / \textbf{98.0} / \underline{88.3} & \underline{95.5} / \underline{97.0} / \underline{84.8} & \underline{95.3} / \underline{97.4} / \underline{84.4} & \underline{95.9} / \textbf{97.5} / \underline{86.1} \\
        \hline
        CLIP & 93.8 / 94.7 / 79.8 & 95.1 / 96.3 / 83.5 & 93.3 / 94.2 / 77.2 & 93.2 / 94.3 / 76.9 & 93.8 / 94.9 / 79.4 \\
        SegFormer & \textbf{96.1} / \textbf{97.5} / \textbf{87.1} & \textbf{96.6} / \textbf{98.0} / \textbf{88.7} & \textbf{95.7} / \textbf{97.1} / \textbf{85.6} & \textbf{95.8} / \textbf{97.6} / \textbf{86.5} & \textbf{96.0} / \textbf{97.5} / \textbf{87.0} \\
        DINOv3 & 95.4 / 97.0 / 84.9 & 96.1 / 97.8 / 87.2 & 94.8 / 96.5 / 83.0 & 95.1 / 97.0 / 84.1 & 95.4 / 97.1 / 84.8 \\
        \bottomrule
    \end{tabular}
\end{table*}

\begin{table*}[t]
    \centering
    \caption{Segmentation evaluation results following Protocol 1. We report the Dice Score and the mean Intersection over Union (mIoU). The best results are highlighted in \textbf{bold}, and the second-best are \underline{underlined}.}
    \label{tab:seg}
    
    \begin{tabular}{l|c|c|c|c|c}
        \toprule
        \multirow{2}{*}{Method} & Qwen-Image-Edit & Flux-Kontext & Step1X-Edit & BAGEL & Average \\
        \cline{2-6}
         & Dice / mIoU & Dice / mIoU & Dice / mIoU & Dice / mIoU & Dice / mIoU \\
        \hline
        ObjectFormer & 68.1 / 61.0 & 75.0 / 67.8 & 63.8 / 56.6 & 61.7 / 54.6 & 67.1 / 60.0 \\
        PSCC-Net & 70.9 / 61.8 & 77.3 / 68.8 & 66.8 / 57.6 & 68.3 / 59.3 & 70.8 / 61.9 \\
        IML-ViT & 72.9 / 67.0 & 82.9 / 77.4 & 70.3 / 64.2 & 71.8 / 65.6 & 74.5 / 68.6 \\
        PIM-Net & 79.0 / 72.2 & 85.2 / 79.4 & 75.2 / 68.4 & 78.1 / 71.4 & 79.4 / 72.9 \\
        Mesorch & \textbf{83.1} / \textbf{78.0} & \underline{86.7} / \textbf{81.8} & \underline{79.8} / \underline{74.3} & \underline{78.7} / \underline{73.0} & \underline{82.1} / \underline{76.8} \\
        \hline
        CLIP & 71.5 / 64.5 & 76.6 / 69.4 & 67.4 / 60.4 & 63.2 / 56.2 & 69.7 / 62.6 \\
        SegFormer & \underline{82.8} / \underline{77.6} & \textbf{87.0} / \textbf{81.8} & \textbf{80.4} / \textbf{74.8} & \textbf{80.3} / \textbf{74.3} & \textbf{82.6} / \textbf{77.1} \\
        DINOv3 & 78.4 / 71.8 & 83.6 / 77.0 & 75.1 / 68.3 & 74.7 / 67.6 & 78.0 / 71.2 \\
        \bottomrule
    \end{tabular}
\end{table*}

\subsection{Baselines}

To establish a comprehensive benchmark, we compare our approach against a suite of state-of-the-art image manipulation localization methods and representative vanilla segmentation models.
The selected manipulation localization baselines include ObjectFormer \citep{objectformer30}, PSCC-Net \citep{psccnet31}, IML-ViT \citep{imlvit32}, PIM-Net \citep{pimnet33}, and Mesorch \citep{mesorch34}. 
For these methods, input images are resized to the recommended resolution of each respective model, and all other configurations follow the original implementations to ensure optimal performance. 
A consistent fine-tuning strategy is applied to the segmentation models, including CLIP \citep{clip37}, SegFormer \citep{segformer36}, and DINOv3 \citep{dinov3_35}. 
This strategy employs a linear head for segmentation across all models, with LoRA \citep{lora38} applied exclusively to DINOv3 for lightweight adaptation. 
These vanilla segmentation models are optimized using AdamW with a fixed learning rate of $1e-4$, trained for $30$ epochs with bfloat$16$ mixed precision. 
To ensure fair comparison under controlled conditions, only horizontal flipping is applied for data augmentation. 
Our method is implemented with the PyTorch library and all the experiments are conducted on $4$ A100 with $40$ GB memory. 

\subsection{Full-Set Evaluation}

We conduct a comprehensive evaluation of existing localization methods on the full training set of the LocateEdit-Bench dataset. 
As shown in \cref{tab:cls} and \cref{tab:seg}, all models achieve non-trivial performance across different editing models, confirming that localizing instruction-based edits is a feasible yet challenging task. 
Pixel-level evaluation measures the alignment of predictions with the ground truth across the entire image, while segmentation evaluation specifically measures this alignment within the edited regions. 
A closer examination of the segmentation scores highlights a critical limitation of traditional pixel-wise metrics: widely adopted metrics like Accuracy and AUC are heavily dominated by the large, untouched background regions, thereby failing to accurately reflect a model's true capability to localize the manipulated areas. 
The overall segmentation performance remains substantially limited, revealing a significant gap between current forensic tools and the requirements for reliably localizing forgeries generated by modern, instruction-driven models.

\begin{figure*}[t]
    \centering
    \includegraphics[width=1.0\linewidth]{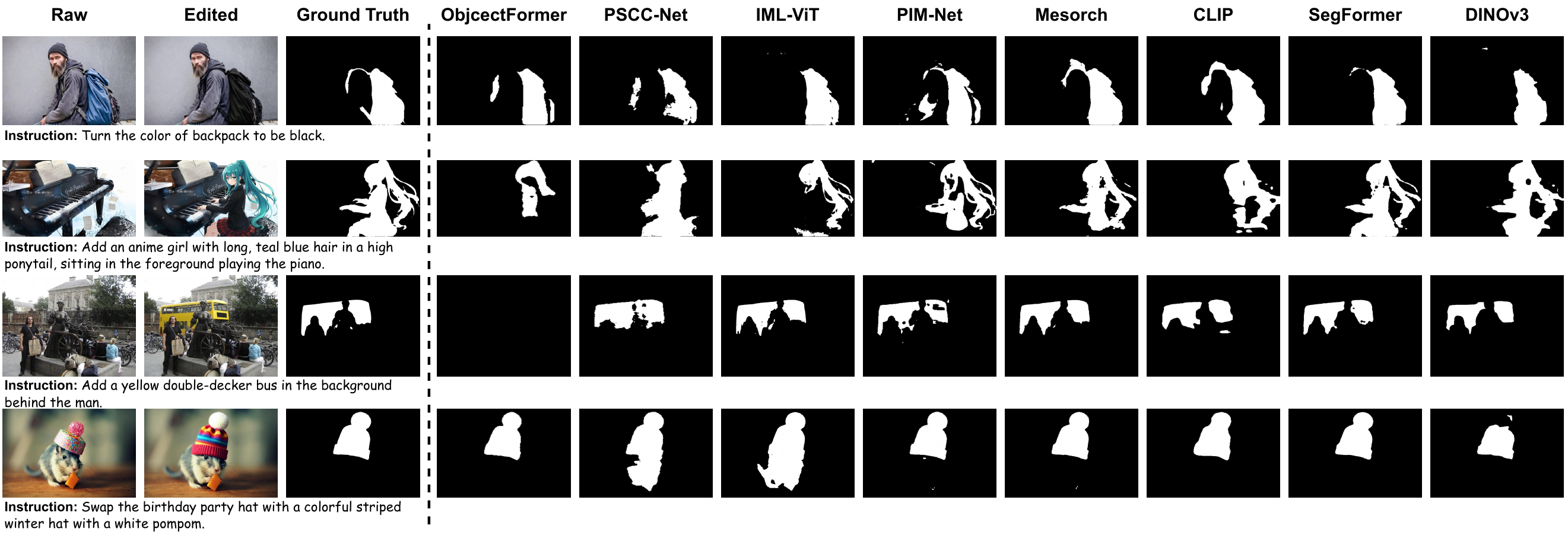}
    \caption{Visualization results of the localization models on the LocateEdit-Bench dataset. }
    \label{fig:visual}
\end{figure*}

\begin{figure*}[t]
    \centering
    \hfill
    \subfloat[ObjectFormer]{\includegraphics[width=0.23\linewidth]{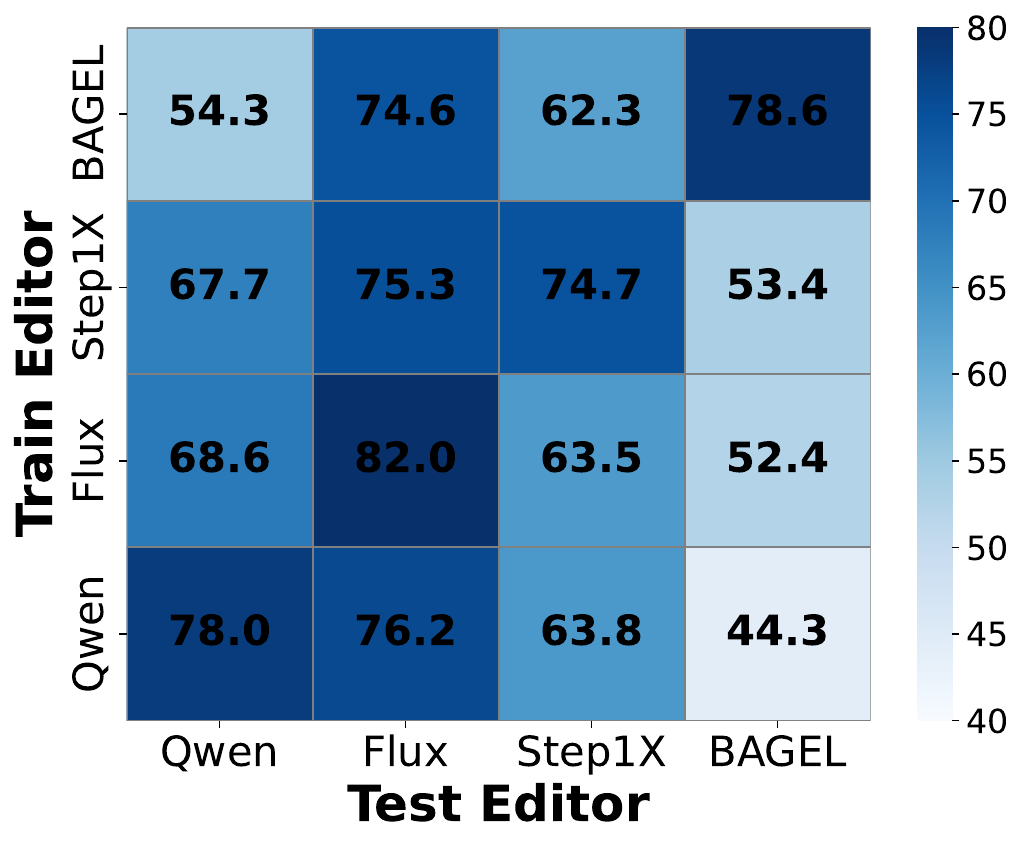}}
    \hfill
    \subfloat[PSCC-Net]{\includegraphics[width=0.23\linewidth]{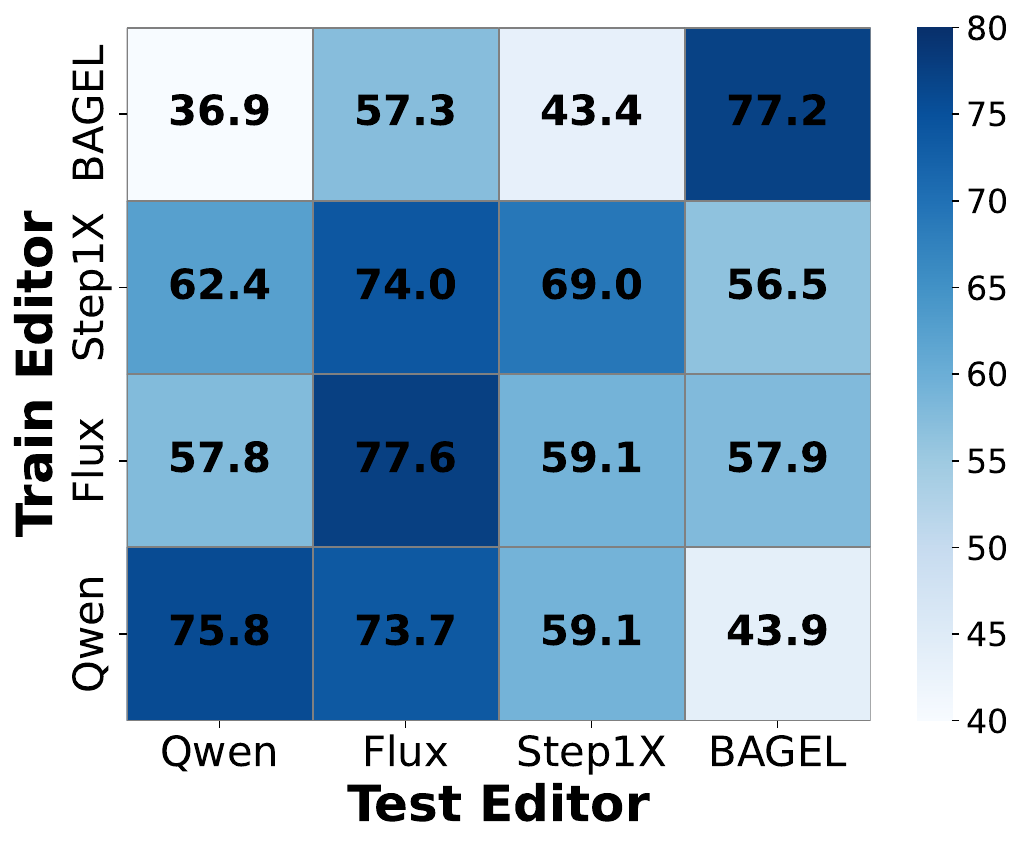}}
    \hfill
    \subfloat[IML-ViT]{\includegraphics[width=0.23\linewidth]{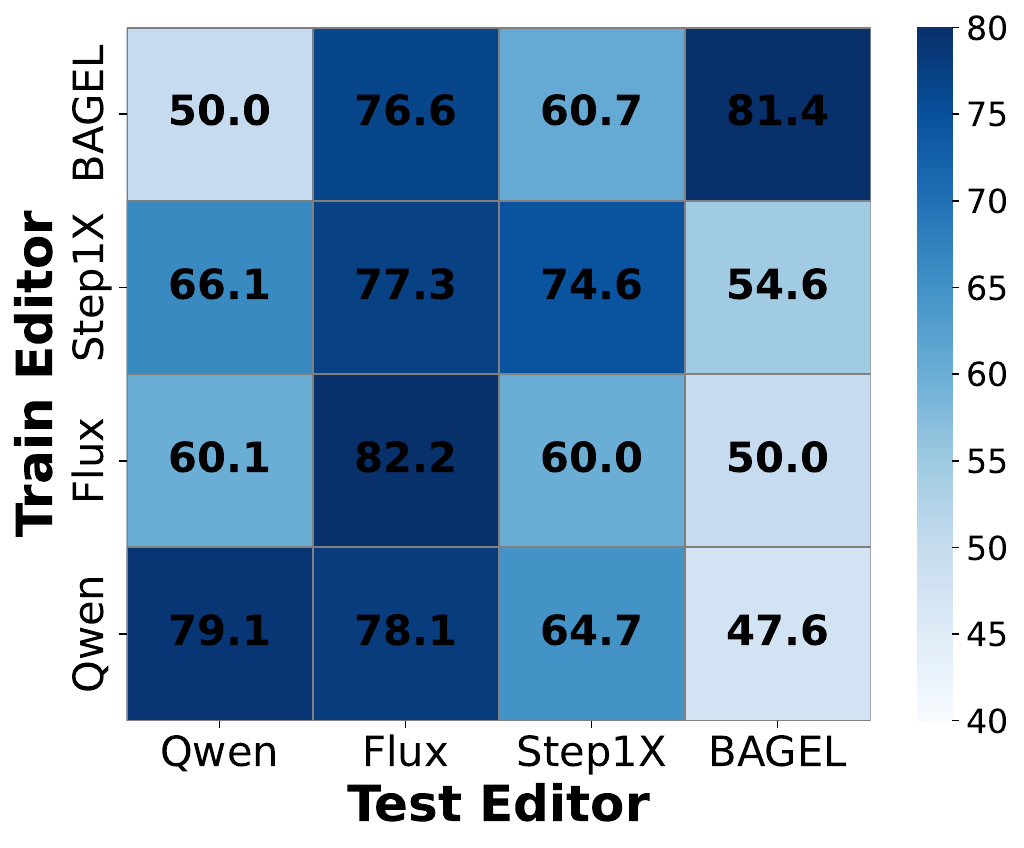}}
    \hfill
    \subfloat[PIM-Net]{\includegraphics[width=0.23\linewidth]{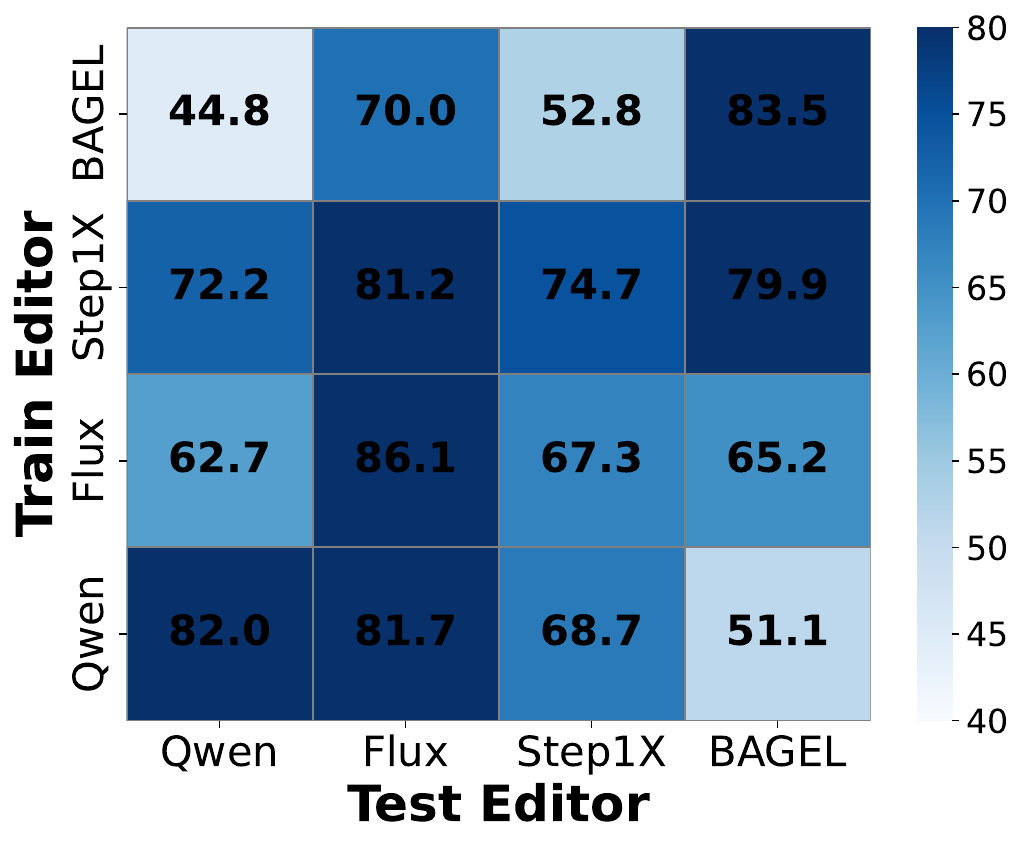}}
    \hfill
    \subfloat[Mesorch]{\includegraphics[width=0.24\linewidth]{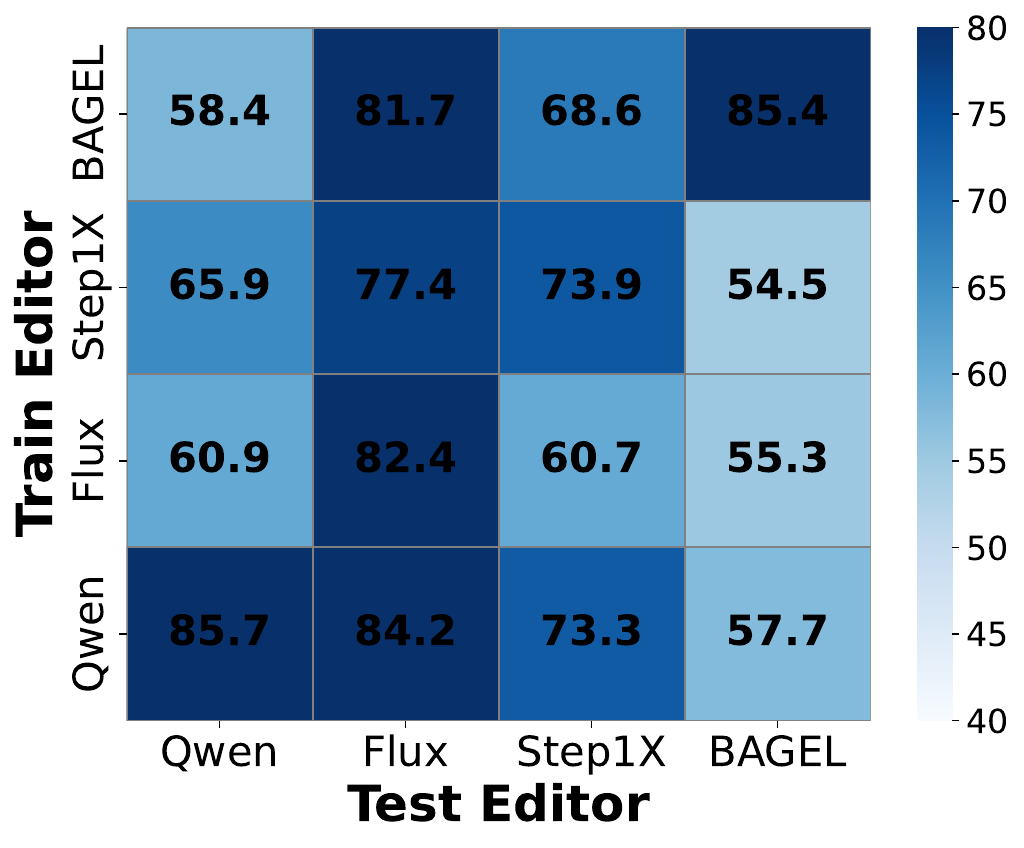}}
    \hfill
    \subfloat[CLIP]{\includegraphics[width=0.24\linewidth]{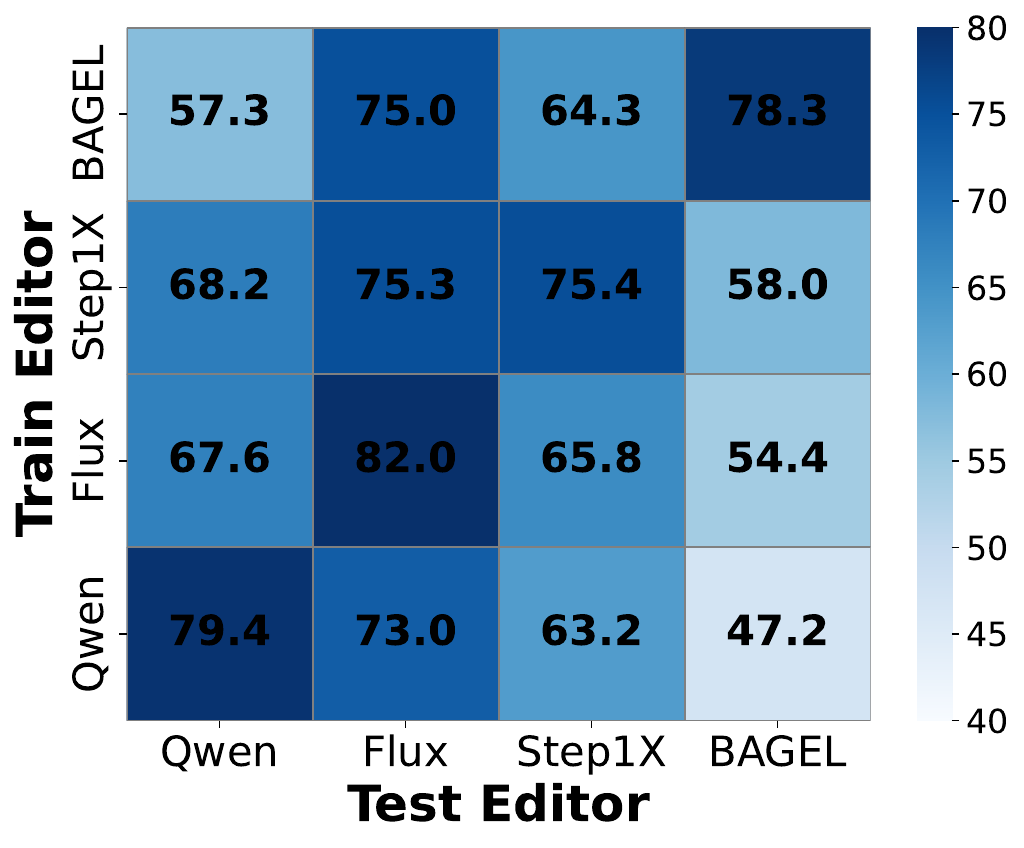}}
    \hfill
    \subfloat[SegFormer]{\includegraphics[width=0.24\linewidth]{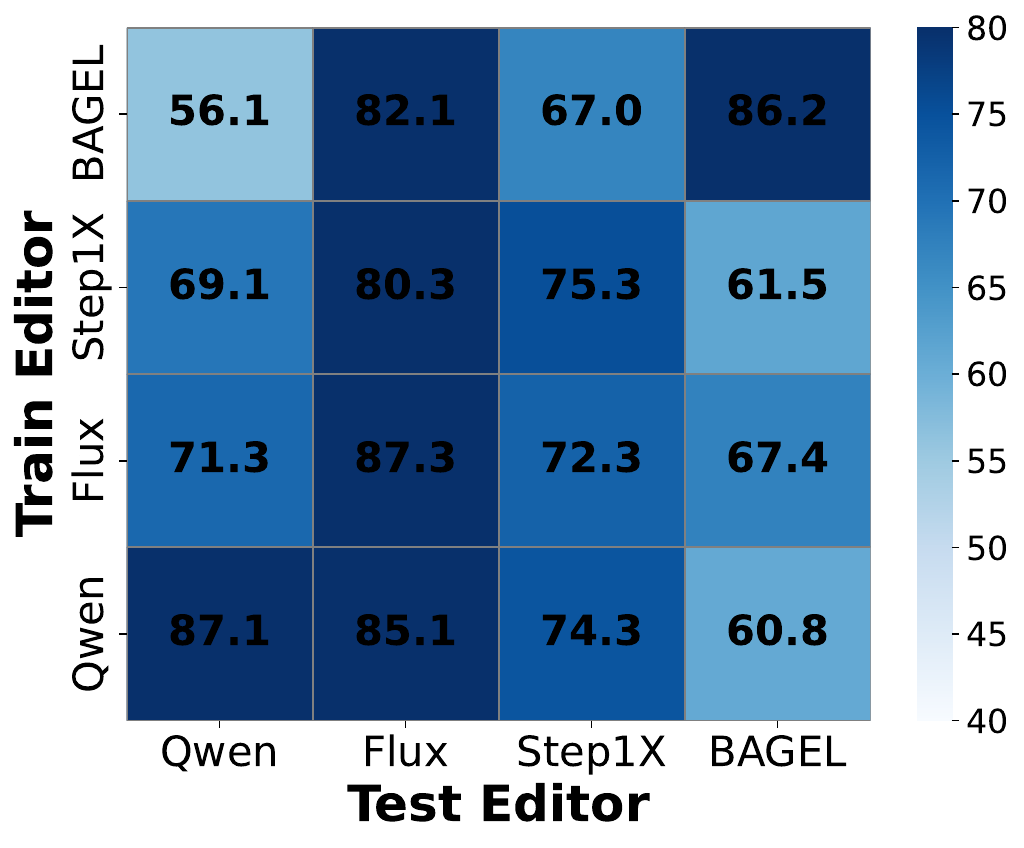}}
    \hfill
    \subfloat[DINOv3]{\includegraphics[width=0.24\linewidth]{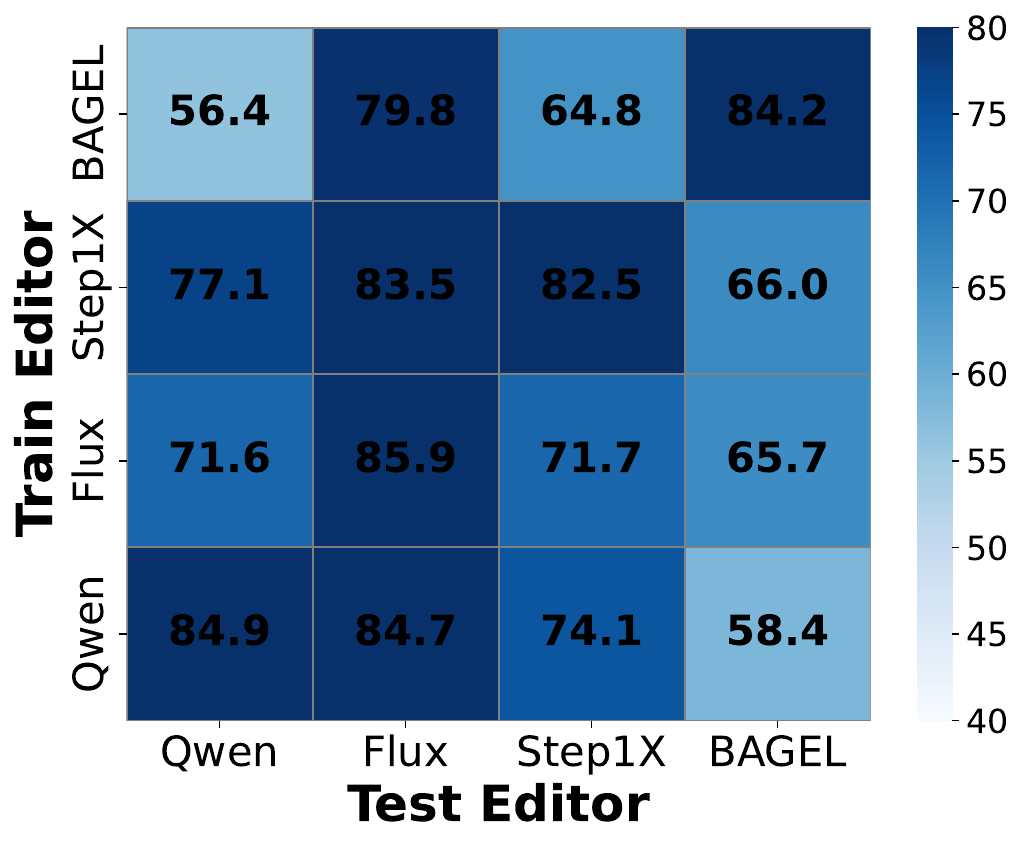}}
    \hfill
    \caption{Cross editor evaluation of the localization models using protocol 2. We report pixel-level F1-Scores to comprehensively evaluate the performance of each model. Most models exhibit a performance drop on editor categories not encountered during training. }
    \label{fig:crossvalid}
    \vskip -0.1in
\end{figure*}

Further analysis reveals a distinct performance hierarchy: SegFormer, a vanilla segmentation model, achieves the best overall performance. 
Its superiority can be attributed to its multi-scale feature fusion mechanism, which effectively captures semantic inconsistencies and editing traces across different spatial resolutions. 
Notably, Mesorch utilizes SegFormer as its backbone and achieves the second-best performance. 
This strongly indicates that deep integration of semantic features is crucial for the task of instruction-based image editing localization. 
In contrast, methods focusing primarily on low-level anomaly detection (e.g., ObjectFormer, PSCC-Net) show significantly poorer segmentation results on our challenge benchmark. 
These findings underscore that effective localization must extend beyond low-level signal analysis to incorporate high-level semantic reasoning. 

The visual results are presented in \cref{fig:visual}. 
Notably, SegFormer and PIM-Net are capable of generating precise, detail-aware segmentation masks. 
In contrast, most other models produce only coarse, approximate regions. 
This indicates that the former two models operate with a deeper, more precise semantic understanding. 
The qualitative observations are consistent with the reported quantitative metrics. 
Furthermore, performance varies noticeably across different editing models. 
The results show that images edited by BAGEL are the most challenging to localize, in contrast to those edited by Flux-Kontext, which are the easiest. 
This indicates that the characteristics of the forgery are highly editor-dependent and often specific to the editing task. 

\subsection{Cross-Editor Generalization}

To assess the generalization capability of localization models when encountering unseen editing algorithms, we conduct a cross-editor evaluation following Protocol 2.
This setup simulates real-world scenarios where a forensic tool must handle manipulations from previously unknown editing models. 
The results are presented in \cref{fig:crossvalid}, which visualizes the F1-Score performance in a cross-editor setting. 
The y-axis of each heatmap represents the editing model type used for training, while the x-axis indicates the model type used for testing. 

\begin{table}[t]
    \centering
    \caption{Robustness to JPEG compression with quality=0.75. Following Protocol 1, the average scores across all four editor types are reported. }
    \label{tab:jpeg}
    
    \begin{tabular}{l|c|c}
        \toprule
        \multirow{2}{*}{Method} & \multicolumn{2}{c}{Average} \\
        \cline{2-3}
         & ACC / AUC / F1 & Dice / mIoU \\
        \hline
        ObjectFormer & 93.3 / 94.4 / 74.9 & 62.7 / 55.7 \\
        PSCC-Net & 89.7 / 93.4 / 70.9 & 66.4 / 57.3 \\
        IML-ViT & 92.7 / 94.2 / 74.2 & 67.0 / 61.1 \\
        PIM-Net & 92.9 / 94.4 / 77.8 & 74.5 / 67.6 \\
        Mesorch & \underline{94.5} / \underline{96.1} / \underline{80.6} & \underline{76.0} / \underline{70.7} \\
        \hline
        CLIP & 93.3 / 94.2 / 76.8 & 67.2 / 60.3 \\
        SegFormer & \textbf{95.0} / \textbf{96.5} / \textbf{82.9} & \textbf{77.9} / \textbf{72.4} \\
        DINOv3 & 94.2 / 96.0 / 79.9 & 72.9 / 66.1 \\
        \bottomrule
    \end{tabular}
\end{table}

\begin{table}[t]
    \centering
    \caption{Robustness to Gaussian blurring with $\sigma$=1.0. Following Protocol 1, the average scores across all four editor types are reported. }
    \label{tab:gaussian}
    
    \begin{tabular}{l|c|c}
        \toprule
        \multirow{2}{*}{Method} & \multicolumn{2}{c}{Average} \\
        \cline{2-3}
         & ACC / AUC / F1 & Dice / mIoU \\
        \hline
        ObjectFormer & 91.5 / 91.0 / 71.7 & 59.6 / 52.4 \\
        PSCC-Net & 71.8 / 86.2 / 50.3 & 52.2 / 42.7 \\
        IML-ViT & 87.3 / 90.1 / 67.9 & 63.2 / 57.8 \\
        PIM-Net & 89.1 / 91.8 / 71.0 & 70.1 / 63.0 \\
        Mesorch & \underline{94.8} / \textbf{96.1} / \underline{82.9} & \underline{78.2} / \textbf{72.7} \\
        \hline
        CLIP & 92.0 / 92.2 / 74.8 & 66.0 / 59.2 \\
        SegFormer & \textbf{94.9} / \textbf{96.1} / \textbf{83.3} & \textbf{78.3} / \underline{72.6} \\
        DINOv3 & 94.2 / 95.5 / 81.4 & 74.6 / 67.9 \\
        \bottomrule
    \end{tabular}
\end{table}

Specifically, all models suffer a substantial degradation in performance when evaluated on unseen editors. 
This indicates that different instruction-based editors may introduce distinct visual artifacts, texture patterns, or semantic inconsistencies even when performing similar editing tasks. 
Consequently, localization models trained on data from a limited set of editors fail to capture these cross-model commonalities effectively. 
Notably, BAGEL-edited images consistently represent the most challenging generalization case.
This can be attributed to BAGEL's unique architecture as a unified generative-understanding model, which aligns semantic instructions with visual content more coherently than task‑specialized editors, thereby producing edits that exhibit fewer typical manipulation traces and higher semantic fidelity. 
Our study highlights this generalization gap, suggesting that the key to robust detection lies in learning editor-agnostic forensic features and establishing frameworks for modeling semantic consistency across diverse editing paradigms. 

\subsection{Robustness Analysis}

This section analyzes the robustness of localization models under real-world conditions, where manipulated images may undergo common post-processing. We incorporate two prevalent types of image degradation: JPEG compression and Gaussian blurring. 
The results, presented in \cref{tab:jpeg} and \cref{tab:gaussian}, show a consistent decline across all evaluation metrics when models process degraded inputs. In particular, the F1-Score exhibits the most pronounced drop, underscoring the particular sensitivity of segmentation-based localization methods to deterioration in image quality. This performance decline suggests that such degradations introduce noise and obscure fine-grained details, making it more difficult for models to distinguish genuine manipulation traces from artifacts induced by the processing itself. 
Methods like Mesorch and SegFormer, which rely heavily on multi-scale semantic features, are superior under pristine conditions. However, they are also more susceptible to distortions like blurring and compression, which directly degrade the clarity and discriminability of features across scales. 


These findings underscore a critical vulnerability in current localization approaches: their performance is closely tied to input image quality. In practical forensic scenarios, manipulated images are often distributed in compressed or blurred formats, which may significantly undermine detection reliability. Future work should therefore aim to enhance robustness against such degradations, potentially through training with augmented data encompassing diverse degradation types or by developing feature extractors more resilient to quality variations.
\section{Conclusion}

In this work, we systematically investigate the emerging challenge of localizing forgeries in images edited by modern instruction-based generative models. 
We introduce LocateEdit-Bench, the first large-scale benchmark dataset specifically designed for evaluating the localization of instruction-based image edits. 
Comprising $231$K edited images generated by four state-of-the-art editors across three fundamental manipulation types, our dataset provides a diverse and challenging testbed for forensic research. 
Through extensive benchmarking of existing localization methods, we show that current techniques suffer from significant performance degradation when confronted with instruction-based edits. 
The cross-editor generalization experiments further reveal the limited adaptability of these methods across different generative models, highlighting the complex nature of this forgery paradigm. 
By investigating this evolving landscape, our work establishes a crucial foundation to facilitate future research towards generalizable and semantically-aware localization in advanced AI-edited images. 



\section*{Impact Statement}

The rapid advancement of instruction-based image editing technologies has enabled the creation of highly realistic and semantically coherent visual forgeries, raising significant concerns regarding digital media authenticity and trust. 
Our work addresses a critical gap in the field of image forensics by introducing the first large-scale benchmark, LocateEdit-Bench, dedicated to localizing such edits. 
This research contributes to the development of reliable detection tools that can help mitigate the spread of manipulated media in contexts such as misinformation, evidence tampering, and digital content verification.
All data used in this study are generated under controlled settings for research purposes. 
We believe our benchmark will support the creation of more robust forensic methods, promoting transparency and accountability in the era of AI-powered image synthesis. 



\bibliography{example_paper}
\bibliographystyle{icml2026}




\end{document}